\def\year{2023}\relax
\documentclass[letterpaper]{article} % DO NOT CHANGE THIS
\usepackage{aaai23}  % DO NOT CHANGE THIS
\nocopyright
\usepackage{times}  % DO NOT CHANGE THIS
\usepackage{helvet}  % DO NOT CHANGE THIS
\usepackage{courier}  % DO NOT CHANGE THIS
\usepackage[hyphens]{url}  % DO NOT CHANGE THIS
\usepackage{graphicx} % DO NOT CHANGE THIS
\usepackage{natbib}  % DO NOT CHANGE THIS AND DO NOT ADD ANY OPTIONS TO IT
\usepackage{caption} % DO NOT CHANGE THIS AND DO NOT ADD ANY OPTIONS TO IT
\DeclareCaptionStyle{ruled}{labelfont=normalfont,labelsep=colon,strut=off} % DO NOT CHANGE THIS
\usepackage{color,soul}
\usepackage[rightcaption]{sidecap}
\sidecaptionvpos{figure}{c}
\usepackage{algorithm}
\usepackage{algorithmic}
\usepackage{newfloat}
\usepackage{listings}
\usepackage{color}
\floatstyle{ruled}
\newfloat{listing}{tb}{lst}{}
\floatname{listing}{Listing}
\title{Steps Towards Satisficing Distributed Dynamic Team Trust}
\author{
Edmund R. Hunt,\textsuperscript{\rm 1} Chris Baber,\textsuperscript{\rm 2} Mehdi Sobhani,\textsuperscript{\rm 1} Sanja Milivojevic,\textsuperscript{\rm 3} Sagir Yusuf,\textsuperscript{\rm 2}\\ Mirco Musolesi,\textsuperscript{\rm 4} Patrick Waterson,\textsuperscript{\rm 5} and Sally Maynard\textsuperscript{\rm 5}
}
\affiliations{
\textsuperscript{\rm 1}School of Engineering Mathematics and Technology, University of Bristol\\
\textsuperscript{\rm 2}School of Computer Science, University of Birmingham\\
\textsuperscript{\rm 3}Bristol Digital Futures Institute, University of Bristol\\
\textsuperscript{\rm 4}Department of Computer Science, University College London\\ 
\textsuperscript{\rm 5}Human Factors and Complex Systems Group, Loughborough University\\

edmund.hunt@bristol.ac.uk, c.baber@bham.ac.uk, mehdi.sobhani@bristol.ac.uk, sanja.milivojevic@bristol.ac.uk, s.m.yusuf@bham.ac.uk, m.musolesi@ucl.ac.uk, p.waterson@lboro.ac.uk, s.e.maynard@lboro.ac.uk
}

\usepackage{bibentry}
\makeatletter
\def\@copyrightspace{\relax}
\makeatother

\begin{document}

\maketitle

\begin{abstract}
Defining and measuring trust in dynamic, multiagent teams is important in a range of contexts, particularly in defense and security domains.  Team members should be trusted to work towards agreed goals and in accordance with shared values. In this paper, our concern is with the definition of goals and values such that it is possible to define ‘trust’ in a way that is interpretable, and hence usable, by both humans and robots.  We argue that the outcome of team activity can be considered in terms of ‘goal’, `individual/team values', and ‘legal principles’. We question whether alignment is possible at the level of `individual/team values', or only at the `goal' and ‘legal principles’  levels. We argue for a set of metrics to define trust in human-robot teams that are interpretable by human or robot team members, and consider an experiment that could demonstrate the notion of `satisficing trust' over the course of a simulated mission.
\end{abstract}

\vspace{-0.5cm}
\section{Introduction}
Defining the interpersonal and technical factors that relate to trust in human-AI/robot teaming is an open problem in the research community \cite{Huang2021}.
In particular a key problem is the definition of `trust' in these scenarios.
%Such an approach agrees with the focus of our own ‘HURST’ (HUman-Robot Satisficing Trust) project.  
We expect trust to vary according to the developing situation faced by each teammate. Thus, obtaining a trust level sufficient for a given situation will always involve satisficing, i.e. obtaining a minimally acceptable level to progress a team’s mission. We use the metaphor of a \textit{Ladder of Trust}, whereby teammates may `climb' down or up the ladder to satisfice situation-dependent requirements \cite{Baber2023}. 

%\section{Characterizing Trust}
To study trust in human-robot teams, it is necessary to define the concept of trust in a manner which is meaningful for both humans and robots. Previously, trust has been seen as a multi-dimensional concept that focuses on human perceptions \cite{McAllister1995,Mayer1995}, e.g., through self-report questionnaires for the human team members \cite{Jian2000,Schaefer2016,Malle2021}, but this assumes cognitive and cultural capabilities beyond those of robots. While humanlike robots can be perceived to be capable of making `moral' decisions \cite{Malle2021}, in general, people do not accept the idea of machines making moral decisions \cite{Bigman2018}. We believe that there has been less attention given to trust held by robots of their human team mates. In our project, we seek metrics appropriate for both humans and robots to quantify, and hence regulate, their trust in their teammates. 

\subsection{Principles, Values and Goals}

\begin{figure}
\centering
        \includegraphics[width=0.25\textwidth]{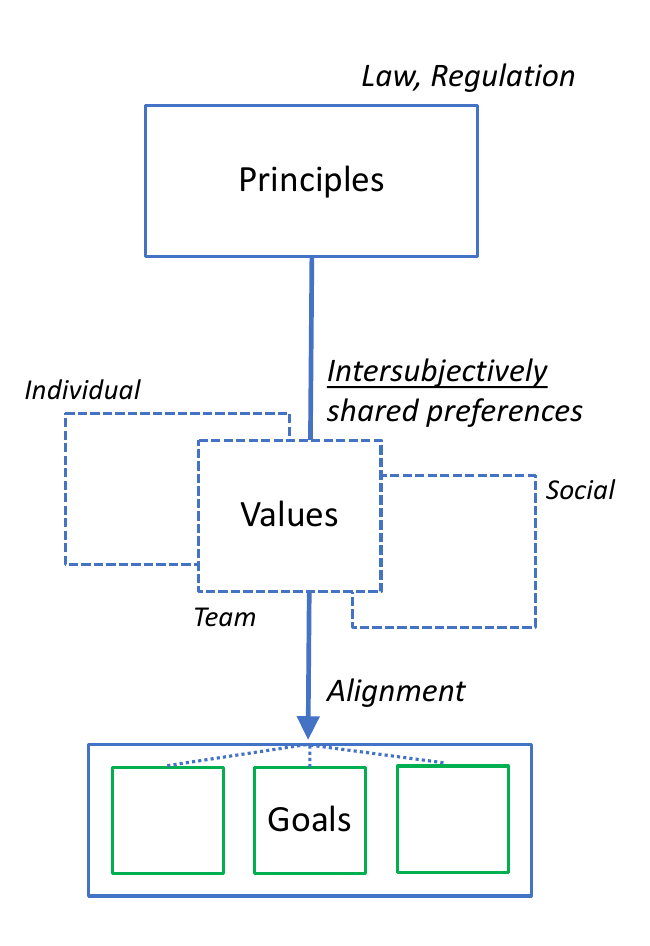} 
    \caption{Alignment of goals and principles between humans and robots is possible; expressing `values' in contextualized goal priorities remains the province of human actors.}
     \label{fig:fig0}
    \vspace{-0.5cm}
\end{figure}

We see a hierarchy of \textit{principles}, \textit{values} and \textit{goals} (Fig.~\ref{fig:fig0}). For human-robot teams, trust requires team members to behave consistently with the team’s goals, team values and legal principles to a standard of performance agreed by the team \cite{Baber2023}. Goals are pre-defined ends, aims or objectives for humans or robots in a team. A team goal, for example, could be to defuse a bomb. Values, on the other hand, have a sense of purpose. They are \textit{`those ends deemed worth pursuing'} (\citeauthor{Williams2009} 2009, p.559). Values define desirable outputs: what motivates us, what is worth striving for. As such, they can be individual, team, and social. Social values, such as liberty, freedom, justice, democracy, and respect for fundamental rights, are accepted by society and are aspirational.  Legal principles are `norms laying down essential elements of a legal order' (van Bogdandy, cited in \citeauthor{Williams2009} 2009, p.559). Following Habermas, we consider legal principles to be general propositions from which norms arise, \textit{`certain standards that might be based in law or practice, which contribute to forming a framework for decision-making and action'} (\citeauthor{Williams2009} 2009, p.559). Acting according to legal principles, in our example, could be to defuse the bomb without a loss of human life. Legal principles have a sense of obligation attached to them; they set the bounds or constraints within which activities are permitted. Values could fill the gap where legal principles fail to provide guidance. For example, acting according to social values would be to defuse a bomb even if that means a loss of human life, where such action saves the lives of many others. Acting according to \textit{team} values would be defusing a bomb with a loss of human life, but with preservation of the capability of the human-robot team to continue its mission. Values are moral in character. They are what is the best for the individual, the team, or society. Principles command; values recommend; goals direct. Principles are binary (valid or invalid), values are not. As such, values can be considered in terms of a trade-off in which individual, team and societal values influence the choice of goal and definition of acceptable outcome in a specific situation.

\subsection{Values in Human-Robot Teams}

To speak of ‘values’ in a human-robot team might imply that these can be defined with sufficient clarity that they can be quantified.  Indeed, the Artificial Intelligence (AI) literature has discussed ‘value alignment’ \cite{Gabriel2021}, through which the ‘values’ that an AI system pursues match those of the human stakeholders affected by the system \cite{Moor2006}. An early example of explicit ethics in AI systems can be seen in the ‘ethical governor’ \cite{Arkin2009}. This was proposed for a weapons targeting system in which selection of target could be overruled if there were rules which the decision violated. In this instance, the `rules' could be defined in a similar manner to our notion of legal principles. We might be able to codify legal principles, as legal norms setting the essential elements of the legal order. But ‘values’ are, as Habermas (cited in Williams 2009) suggests, \textit{intersubjectively shared preferences}. The concepts of `value alignment' and `explicit ethics' in AI seem to rest on an assumption that it is possible to codify ‘social values’ with sufficient clarity that an action might be evaluated. We are skeptical that this might be possible.  We believe that alignment of goals and principles (between humans and robots in a system) is possible, but that ‘social values’ must remain the province of the human actors who are either directly participating in the system or acting as overseers, regulators, or other stakeholders.  Our emphasis on context could make such an assumption difficult to articulate (because the ‘values’ might vary with context).   We note that approaches that use, for example, an ‘ethical consequence engine’ \cite{Winfield2014}, might deal with context through the simulation of the outcome of an action.  However, this still implies the codification of ‘values’ within a normative moral framework. Nonetheless, while our position is that ‘values’ remain in the human domain, we accept that it might be possible for engineered systems to seek ‘\textit{goal} alignment’ (i.e. between humans and robots).    

The interpretation of a goal (in terms of the desirability of its outcome) will depend on the context in which it is performed.   From this perspective, we might define the problem as one of inferring goals from the activity of other teammates as a problem of Inverse Reinforcement Learning \cite{Ng2000}.   In a recent study, the tension between selfish and social choices (as a basic version of moral problems in social dilemma games) was explored using Reinforcement Learning \cite{Tennant2023}. In this study, Reinforcement Learning agents are provided with intrinsic rewards that reflect different views of ethics (i.e., utilitarian, deontological, consequentialist) and play a variety of iterated social dilemma games (i.e., Prisoner’s Dilemma, Volunteer’s Dilemma, Stag Hunt).  Within a game, an agent seeks to respond to the state of the game by performing an action that will maximize both the game reward and an intrinsic \textit{moral} reward.  Pitching agents with different  moral stances against each other revealed systematic differences in strategy, particularly in terms of cooperative or exploitative activity. Alternative strategies or stances can be considered in terms of counterfactual analysis, which is a fundamental element of causality analysis \cite{pearl2009causality, pearl2018book}, and can be used to understand which changes to a particular data model would change the model decision. Counterfactual analysis in a multi-agent reinforcement learning environment \cite{forney2017counterfactual} could support performance in mission-critical environments through the ability to review alternative courses of action. We regard goals as explicit statements of intent that team members can choose to pursue, and also recognize their pursual by others.  Thus, if an actor (human or robot) sees a teammate performing a sequence of tasks, it might be reasonable for it to assume that this sequence is directed toward achieving a specific goal, and that an alternative goal might be more desirable (to achieve particular team or social values) in that context. This area has been of great interest for the community in the recent years \cite{pmlr-v97-jaques19a,conitzer2023foundations}. 

\section{Trust Specification}

To define `trust', we follow \citet{Lewis2022} in claiming that a minimal specification of trust involves:
\begin{enumerate}
    \item \textit{Capability}, i.e., is the teammate most appropriate for a given task in that situation?
    \item \textit{Predictability}, i.e., is the teammate acting in a way that fits the team goals and set principles, and is appropriate to its situation?
    \item \textit{Integrity}, i.e., is the teammate acting to support the team and the set principles? 
\end{enumerate}

\begin{figure}[t]
\centering
\includegraphics[width=0.9\columnwidth]{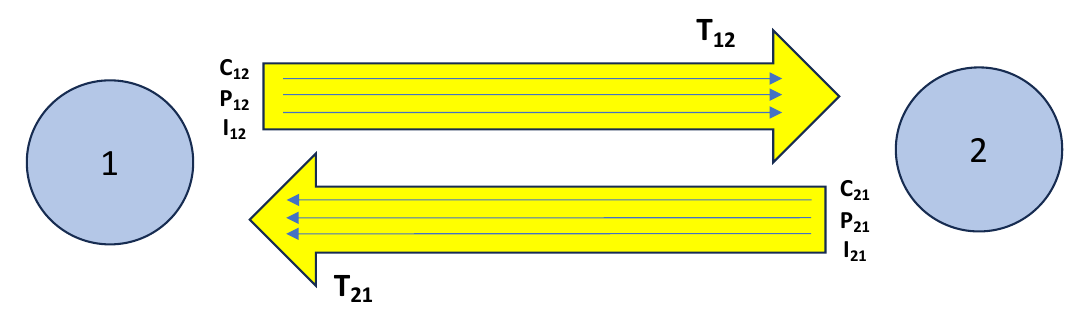} 
\caption{Agent 1’s trust in Agent 2, $T_{12}$, depends on its running estimates of Agent 2’s capability, predictability and integrity; these are contextualized by the local situation.}
\label{fig:fig1}
\end{figure}

In order to define metrics for the three-element view of trust outlined above, we consider what could be sensed or perceived when humans and robots interact in a collaborative task. The literature on trust generally refers to dyadic relationships between a ‘trustor’ and ‘trustee’ (e.g. \citealt{Hurley2006,Kim2009}) and this can be represented as a network of directed edges, where each edge represents one of the three elements (Figure~\ref{fig:fig1}). We do not suppose that the three elements can be measured with equal certainty (some aspects might need to be inferred rather than perceived), but each agent will track the relative increase or decrease of these elements over time to adjust their `trust' in a teammate. 

Recent work has introduced the terms system-wide trust and component-level trust \cite{Walliser2023}.  These map well to our view of distributed, dynamic team trust, where we would clarify component-level trust as comprising the three elements of estimated capability, predictability and integrity, as contextualized by the agents’ situation. As \citet{Huang2021} note, it is important to understand the context in which human-AI-robot teaming occurs “including the tasks, environment, the stakeholders, and artificial agents involved…[as well as]…the kinds of interactions that are available between the entities involved in a situation context, where transitive properties of trust take place.” (\citeauthor{Huang2021}, p.307).  From this, \citet{Huang2021} propose a 4-step process that involves identification of context and stakeholders and defining and measuring trust relationships in the trust network.  In our approach, we define a mission using Cognitive Work Analysis (CWA) (\citet{Rasmussen1994,Vicente1999,Jenkins2009}).

From the Work Domain Analysis phase of CWA, a mission can be decomposed into an Abstraction Hierarchy that (reading from top-to-bottom) describes the \textit{purpose} of the system and (reading from bottom-to-top) describes the \textit{activity} of the system. The claim is that any system is intended to achieve an outcome (or set of outcomes) that can be evaluated in terms of desirable consequences.  Such consequences reflect the values of the stakeholders working with and affected/impacted by the system and can serve as constraints on the goals that the system is seeking to achieve. Goals could be defined by more than one `value', and the values might conflict. Where there is conflict, this either requires negotiation between teammates or intervention by a higher authority.  Legal principles could, to some extent, define the constraints within which a mission is performed, but these will need to be filtered through values appropriate to the situation.

With our metaphor of a `ladder of trust’ \cite{Baber2023}, it is these sub-component-level trust metrics (e.g. $C_{12}$, $P_{12}$, $I_{12}$, indicating Agent 1's estimation of capability, predictability, integrity of Agent 2 in Fig.~\ref{fig:fig1}) that will change as more information is gathered from their interactions. To satisfice trust, then, means for the necessary combination of component (agent) level trust estimates, each comprising these three elements, to reach a certain threshold before a mission can proceed effectively. In practical terms, this might mean Agent 1 improving its estimate of Agent 2’s integrity ($I_{12}$, an element of $T_{12}$) before it believes that Agent 2 will carry out a certain task within the mission in an acceptable manner. As a working hypothesis, we assume that system-wide trust is limited by its weakest component. In cases where an action only requires a subset of team members to be carried out, deficiencies in system-wide trust need not hold back progress, but instead the demand is to satisfice the relevant combination of component-level trust between the situated team members.

\begin{figure}[t]
\centering
\includegraphics[width=0.62\columnwidth]{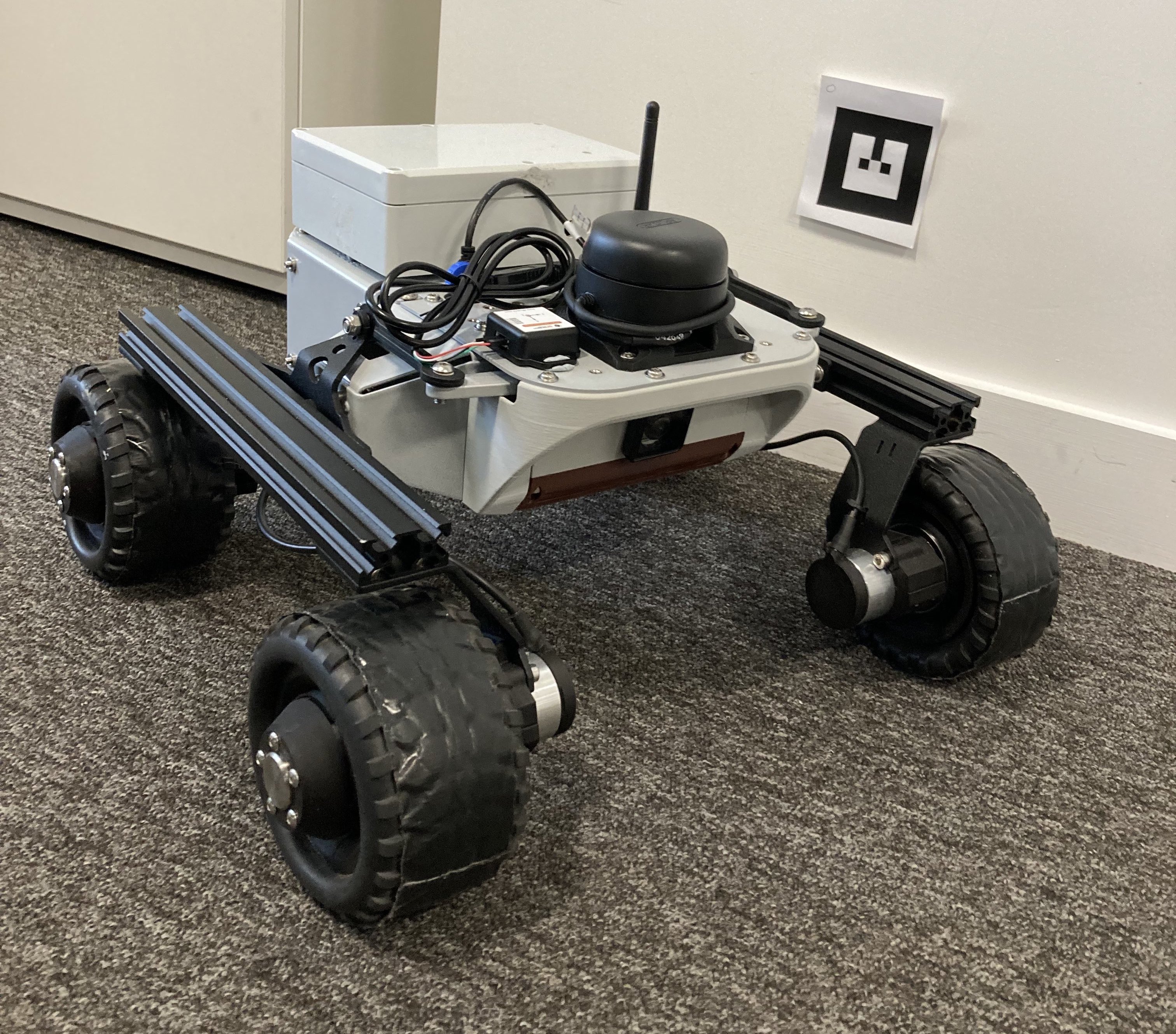}
\caption{A ‘Leo Rover’ robot equipped with 2D LiDAR and built-in front facing camera, which can scan ARTags.}
\label{fig:fig2}
\end{figure}

\section{Motivating Use-Case}
We have designed an environment in which the concept developed in this paper can be experimented upon.  Human participants work with wheeled rover robots (Figure~\ref{fig:fig2}) to collaborate on search tasks in an environment. In the environment, Augmented Reality (AR) Tags are positioned at approximately 1 m height, for the human to scan, and also at approximately 20 cm height, for the robots to scan (Figure~\ref{fig:fig2}). The tags provide information about the environment (whether an area is safe or hazardous to the human) and status of objects in the environment (e.g., whether these are operational, risky, or require repair). For instance, the tags could indicate whether a location has low or high levels of radiation or could indicate whether there is a suspect package that needs to be investigated. The team might comprise two robots and one human in the field and one human working as a coordinator (remotely). Together, the robots and humans need to coordinate a search mission in an experimental environment. As the mission is performed, data can be collected from the human (using a computer tablet interface) or from sensors on the robot. The data from human and robot actors follow the same structure, e.g., \{agent ID, time, location, object, action…\}. All the data and communication between team members (robot(s) and human(s)) are relayed through a Django server and recorded in a database. For instance, the mobile application (‘app’) used as an interface for communicating to the robot is connected to the server so that any commands from a human to a robot will be recorded along with other relevant information like time. The robot messages to human teammates also goes through the server before appearing on the app. The location of all team members are also being published on the server to record the tracking data before being passed to the interface.    

\begin{table*}[h]
\centering
\resizebox{0.8\textwidth}{!}{%
\begin{tabular}{l|l|l}
Category & Variable                                           & Derivation                                              \\ \hline
         & Time at which an action is performed               & Clock reading for logged data                           \\
WHEN     & Prediction of future action to be performed        & \textit{Inferred from mission plan}                    \\
         & History of previous actions                        & Store of logged data                                    \\ \hline
WHERE    & Fixed location                                     & Location of, e.g., AR Tag                                \\
         & Path                                               & Waypoints recorded from, e.g., ultrasonic tracking \\ \hline
WHAT     & Object                                             & Identification of object used                           \\
         & Task                                               & Action performed on an object                           \\ \hline                        
         & Agent                                              & Agent ID                                                \\
WHO      & Goal                                               & \textit{Inferred from mission plan}                     \\
         & Reputation                                         & Store of integrity from prior actions                   \\ \hline
    & Robots: Compatibility of own and other's goal               & \textit{Inferred individual/team goal priorities}                \\
      WHY   & priorities and compliance with codified principles                &   \\
         &Humans: As above, plus judged performance criteria,   &\textit{Inferred from individual, team and social values}     \\
         & constraints, and determinants of acceptable outcomes & 
\end{tabular}
}
\caption{Initial set of metrics to be measured or inferred. Similarly applicable to humans and robots, except for `Why'.}
\label{tab:tab1}
\end{table*}

\section{Defining Metrics for Trust}
In order to interpret a sequence of tasks and their relation to a goal, it is necessary to define metrics through which observation of tasks and inference of goals can be performed by team members. Table 1 indicates the variables that we seek to capture and whether these can be directly measured or inferred. In Table~\ref{tab:tab1}, entries that are in plain text can be directly obtained from the agents, their activity, or the database recorded during the experiment; entries in italic text can be inferred from the mission plan, i.e., there will be goals that are achieved by performing combinations of tasks.  Finally, the ‘why’ will be derived from the competing values of individuals, the team and society. 

Having outlined a set of metrics (Table~\ref{tab:tab1}), we can relate these to the three-element model as follows:

\begin{itemize}
\item Capability (of an Agent) is a function of WHAT (task, object) and WHO (goals). We assume that a given agent will perceive the affordances of the environment in terms of their own and others' ability to perform a task on an object in pursuit of a goal.
\item Predictability (of an Action) is a function of Capability and Context (i.e., WHERE and WHEN). We assume that an agent will perceive a teammate in a context and infer the likelihood of success of an agent (with a perceived capability) achieving an outcome.
\item Integrity (of an Action) is a function of Predictability and WHY.  For \textit{robot actors}, an action is interpreted by a robot and any onlooking robot teammates in terms of the likelihood of success relative to the individual/team \textit{goal priorities} and compliance with codified principles. For \textit{human actors}, an action is also interpreted by the human and any onlooking human teammates in relation to fulfilling relevant goal priorities in compliance with legal principles; \textit{and also} performance criteria, constraints, and determinants of acceptable outcomes, \textit{as influenced by individual, team and social values}. 
\end{itemize}
Agents will begin a mission with initial teammate-specific estimates for each of these elements, which could be based on factors such as prior experiences with the teammates over the longer-term (e.g. \citeauthor{DeVisser2020}, 2020), or proxies for trustworthiness (e.g. \citeauthor{Lewis2022}, 2022). 

Notice that we are claiming that integrity dynamics are linked to each distinct, situated action rather than an agent's identity.  This is a very different perspective to conventional definitions of integrity; in our view, this allows us to attribute changes in integrity without requiring a theory of mind (as in e.g. \citeauthor{DeVisser2020,Mou2020}, 2020).  Our argument is two-fold.  First, integrity arises from the action performed in a context (which we capture in the predictability function) and the individual and team values, as well as set principles by which the action can be judged.  For example, assume that some actions can benefit the agent (‘selfish’) and others benefit the team. In a context in which no other teammate will be affected, a ‘selfish’ action might be judged neutrally, but in a context where the action might be chosen in preference to one which could aid a teammate, the judgment might be negative. Or, in a context where some actions benefit the agent and the team, but are clashing with set principles (‘do not break the law by X’), the judgment will be negative. Second, each agent will have a \textit{reputation} which reflects the history of these instances of integrity. As other teammates learn the history of the actions performed by an agent, so the reputation of that agent will be formed.  Given an understanding of a teammate’s reputation, one can define an expectation of the action that they might perform.  This can be modeled as a reinforcement learning problem \cite{Anastassacos2020, Anastassacos2021}.

\subsection{Integrity, Reputation and Values}

In our definition of trust, each member of a team will form estimates of their teammates' capability, predictability and integrity as task performance is observed. Observation might be of the actual performance of the task or an outcome of this performance (either the result of the task or a report of from another agent). As information is accumulated, an agent will infer the \textit{reputation} of its teammates. Reputation is a quantifiable factor, and (as we noted above) is based on the history of performing tasks in pursuit of goals relative to the individual and team values.  However, rather than the expectation of prosocial behavior being fixed solely by reputation, we assume that this will be moderated by context.  For example, a teammate might have a reputation for performing ‘selfish’ actions, i.e., seeking to gain rewards for themselves at the expense of their teammate. However, in a given situation, there might not be an option to assist a teammate or the outcome of the action might not be detrimental to the team. In this case, the severity of the outcome of a selfish action might be minimal. When there is sufficient information about a task being performed, the integrity of this task can be defined in terms of selfish or team values. When there is insufficient information, knowing the relationship between task and goal, and the context in which the task is performed, can enable the prediction of the most likely goal being pursued. We describe this by updating a Bayesian Belief Network that is held by each agent (Figure~\ref{fig:fig3}). Note, here we tend to think of reputation as being a short-term, mission-bounded metric, but we also recognize that this metric could interact with other factors, e.g. long-term trust weightings such as relationship equity \cite{DeVisser2020}, or short-term weightings such as robot appearance (which may have a proxy trust effect; \citeauthor{Lewis2022}, 2022); such weightings could be added into the BBN.  

\section{Trust and Distributed Situation Awareness}
We assume agents have a partial view of the context.  This view consists of their perception of the environment, their inference of which goal is appropriate to perform in the context, their belief in the reputation of their teammates, and the goal that they expect their teammates to be pursuing.  Situation awareness in a team is likely to be distributed \cite{Stanton2006} and we have previously demonstrated that Distributed Situation Awareness can be formally described using a Bayesian Belief Network (BBN) model \cite{Yusuf2022}.

\begin{figure*}[t]
\centering
\includegraphics[width=0.82\textwidth]{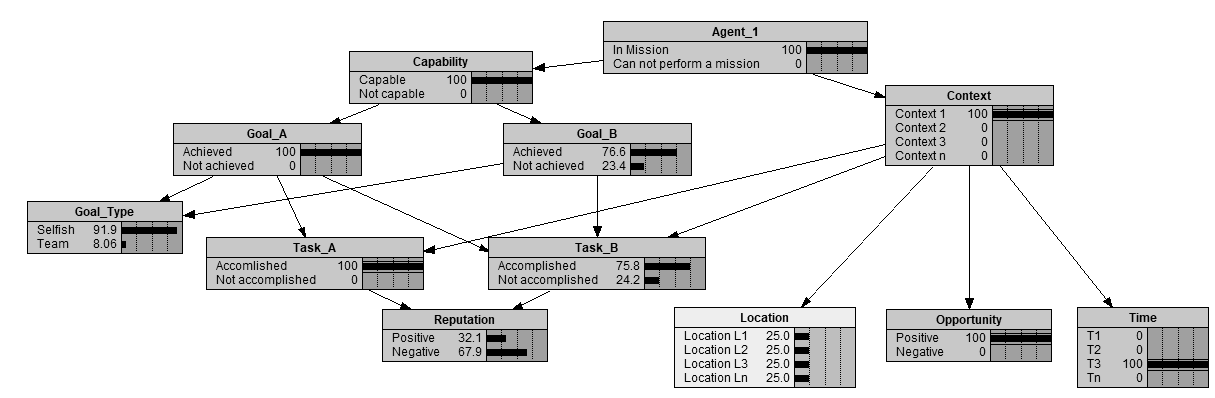}
\caption{A teammate's Bayesian Belief Network (BBN) based on their observation of Agent 1's choice of action in a context.}
\label{fig:fig3}
\end{figure*}

In a BBN, the system can be modeled using a graph $G(N,E)$ where $N$ is a set of nodes connected by a causal directional edges $E$. Each node represents an element of component-level trust with a defined number of states (i.e., selected situations as illustrated in Figure~\ref{fig:fig3}). States are defined as probabilities, i.e., between 0\%-100\% to reflect the assumption that these are estimates formed by an agent. These probabilities can be updated based on mission information (e.g., Equation \ref{eqn:equation1}), or learned over time \cite{Yusuf2022}.  As such, the trust elements (e.g., goals, tasks, reputation etc.) can be modeled using BBN nodes with assigned probabilities and causal relationships based on the operating mission context. For example, assume a scenario where an agent has two goals ($G_A$ and $G_B$), e.g., $G_A$: to mark the locations of hazards in an environment by scanning AR Tags, and $G_B$: to construct the map of the environment using Simultaneous Localization and Mapping algorithms (SLAM).  

Each of these goals can be achieved by completing a number of tasks $\alpha_i, \forall \space i \in [1,N]$; and these tasks can be spread across goals (e.g., the SLAM task to search for a hazard is the same as the one for a mapping task) i.e., $G_A \rightarrow a_i \times a_j$, such that, $\alpha_i \cap \alpha_j \neq \{\},  \forall \alpha_i, \alpha_j \in G_A \cap G_B,  i,j \in [1,N]$ or mapped singly to a goal $G_i \rightarrow \alpha_i, \forall i \in [1,N]$ as illustrated in Figure~\ref{fig:fig3}. From Figure \ref{fig:fig3}, an agent is capable or incapable of achieving a goal, and the goal achievement depends on the assigned task(s) and context (defined by time, location, and opportunity).

Thus, each task achieved will increase the goal’s probability of success, i.e., a task with a 90\% `achieved' state has a higher contribution to the goal achievement than the one with 50\% (though this depends on the criticality of the task towards the goal achievement \citet{Yusuf2022}).  Equation \ref{eqn:equation1} is an example of a protocol-based mode of updating a probability of each state of the BBN after every mission event (e.g., sensor sampling by the agent):
\begin{equation}
\label{eqn:equation1}
P(R_i) = P(R_{i-1}) + f(w_c)
\end{equation}

\noindent where  $P(R_i)$ is the updated prior of the state (e.g., after the event occurrence), $P(R_{i-1})$ is the previous prior of the state,  $f(w_c)$ is the probability decrement/increment weight function (to be assigning values based on subject matter expert (SME) judgments of the context $c$, e.g., information from a reliable sensor weights a complement of 100\%), $c$ is the mission context (e.g., as defined by time, location, and opportunity in Figure \ref{fig:fig3}), and $i$ is the event number, such that, $i \in [1,N]$.  For example, if a goal has two tasks with a weight ratio of 9:1 (e.g., as assigned by $f(w_c)$ of Equation \ref{eqn:equation1}) towards goal achievement, accomplishing task A contributes to $90\%$, i.e., $0\%$ (as the current probability value of the 'achieved' goal state, i.e., assuming no prior progress on task B) $+90\% = 90\%$ for the designated goal achieved state. Note that, $f(w_c)$ reduces the prior $P(R_i)$ for the non-occuring states and sum up the states probabilities to $100\%$.

The probability of a parent node can be defined by the child(ren) 
node(s) states using the conditional probability table (CPT), i.e., a table mapping parent node states probabilities with the joint child(ren) states. One of the advantages of modeling the system concepts using a BBN is the ability to predict states using conditional probabilities (Equation \ref{eqn:equation2}):
\begin{equation}
\label{eqn:equation2}
P(A) = P(A \cap B)/P(A|B)
\end{equation}

\noindent where $P(A)$ is the expected probability of the querying state $A$, $P(A|B)$ is the conditional probability of state A given B, and $P(A \cap B)$ is the joint probability of $A$ and $B$. As such, based on the previous reputation of an agent, its capability on a particular task can be predicted. Expectation maximization algorithms can be used to improve the prediction accuracy, i.e., by checking the agent's performance history \cite{Yusuf2022}.

\section{Illustrating a Ladder of Trust in the Experimental Environment}\label{sec:Exp}

Having defined trust and proposed how it might be measured and captured in a BBN, we now explain how this framework can be tested in our experimental environment. From the motivating use-case, scanning the AR tags to define the safety of an area could be defined as a `team' goal, in that it will benefit other team members. This goal could be further emphasized by giving the robot a competing ‘selfish’ goal: each scan could involve a cost to the robot (e.g., each scan requires energy and the robot will need to leave the environment to recharge when energy level are below a threshold) and the robot could be rewarded for minimizing cost (e.g., by staying as long as possible in the environment).  The human could be rewarded for scanning all AR Tags within a time limit, but there could be a cost to entering unsafe areas.  Scanning an AR Tag in an unsafe area could also represent a cost to robots, who would need to enter that area to help the human (e.g., by guiding the human to safety).  

Defining the potential for conflict between team members allows us to manipulate selfish and team goals, and hence to manipulate the `values' of the team performance.  Each team member will be asked, at set intervals, to rate what their teammates are doing (in terms of expectation of the actions between selfish or team goals). It is possible that the task is at the limits of the agent’s ability or demands significant resources. In this case, the expectation of a trustor is that the teammate (trustee) is exerting themself to achieve the outcome. For a ‘selfish’ goal this could be interpreted as the agent recklessly pursuing a reward at personal cost (leading to an increase in distrust); for a ‘team’ goal this could be interpreted as the agent risking themself for their teammates (leading to an increase in trust across trustors).  As well as expectations about capability, the task will be interpreted in terms of predictability of outcome, i.e., is the outcome likely to be completely successful? As with the previous example, one might expect the reputation assigned to a teammate to be affected by the success of the outcome.  Successful outcome(s) will increase perceptions of that teammate’s capability \cite{Grillo2022}; though if not calibrated could lead to overtrust \cite{ullrich2021development}. The final element, integrity, relates to the interpretation of the goal against the goal priorities (robots and humans) and values (humans only) that the individual or team applies. 

\section{Discussion}
Distributed, Dynamic Team Trust \cite{Huang2021} requires metrics that reflect the activity and interactions of members of a team.  In this paper we share our definition of such metrics and illustrate how these can be applied to the conduct of a mission. 

We contribute to the debate on trust in human-robot teams in the following ways. 
First, we propose that trust arises from a hierarchy of principles, values and goals. Second, we argue that ‘integrity’ (as a component of trust) should be judged in terms of the observed choice of task in a given context; this can lead to an inference regarding the choice of goal in that context.  Where the goal might be considered selfish and where this might have negative consequences for teammates, this will lead to a lower perception of the integrity of the action.  Over a history of observations of such choices made by a specific actor, the ‘reputation’ of that actor will be formed. It is likely that such a reputation will (a) reflect the observations of specific agents and, thus, might differ between agents, but (b) could be shared between agents.  In contexts where there is no history to draw upon and hence, no evidence on which to define a ‘reputation’, this will either have to be assumed by the observing agent (e.g. `proxy trust'), or communicated by the other agent.  For example, a robot might be programmed to assume that its human teammates will behave in a prosocial manner. This would lead it to ascribe a positive reputation to its human teammate – until it collected sufficient evidence (from observation or from other reports) to the contrary. Third, we argue that ‘trust’ is dynamic and context-dependent. In this, we are in agreement with \citet{Huang2021}.  Our approach has been to define the metrics for which each member of a team is able to acquire information. We aim to define these metrics in a way that allow humans or robots to perceive sufficient data to update similar Bayesian Belief Networks (BBN).  This is not to assume that human reasoning is reducible to a BBN but allows the human to infer the robots’ choices (from a BBN that represents the robots’ behavior) and the robot to infer human choices.  Fourth, the concept of a ladder of trust (on which perceptions of teammates can move up and down) provides a metaphor for the ways in which trust changes during a mission.

We intend to develop these concepts in a concrete way by carrying out a form of the experiment described in this paper, to observe whether our operationalization of trust metrics obtains plausible dynamics and supports our notion of ‘satisficing trust’ on a ‘ladder of trust’ \cite{Baber2023}. We hope to obtain insights that can guide both human and robot actors in dynamic trust building.

\section{Acknowledgements} 
The research reported in this paper is supported by grant EP/X028569/1 ‘Satisficing Trust in Human Robot Teams’ from the UK Engineering and Physical Sciences Research Council. This project runs from 2023 to 2026 and involves the Universities of Birmingham, Bristol, Loughborough and UCL.

\bibliography{hurst}

\end{document}